\documentclass[letterpaper, 10 pt, conference]{ieeeconf}  

\IEEEoverridecommandlockouts                              

\usepackage{graphicx}
\usepackage{amsmath} 
\usepackage{amssymb}  
\usepackage{xcolor}%
\newcommand{\revisionAdd}[1]{\textcolor{black}{#1}}

\title{\LARGE \bf
Shape Completion and Real-Time Visualization in Robotic Ultrasound Spine Acquisitions
}

\author{Miruna-Alexandra Gafencu$^{1,2,3*}$, Reem Shaban$^{1 *}$, Yordanka Velikova$^{1,3*}$, \\
Mohammad Farid Azampour$^{1,3}$, and Nassir Navab$^{1}$%
    \thanks{*Shared contribution.}%
    \thanks{$^{1}$ Computer Aided Medical Procedures, Technical University of Munich, Munich, Germany
        {\tt\small email: miruna.gafencu@tum.de}}%
    \thanks{$^{2}$ Konrad Zuse School of Excellence in Reliable AI (relAI), Germany}%
    \thanks{$^{3}$ Munich Center for Machine Learning (MCML). Germany}%
}

\usepackage{booktabs}
\begin{document}

\maketitle
\thispagestyle{empty}
\pagestyle{empty}

\begin{abstract}

Ultrasound (US) imaging is increasingly used in spinal procedures due to its real-time, radiation-free capabilities; however, its effectiveness is hindered by shadowing artifacts that obscure deeper tissue structures. Traditional approaches, such as CT-to-US registration, incorporate anatomical information from preoperative CT scans to guide interventions, but they are limited by complex registration requirements, differences in spine curvature, and the need for recent CT imaging. Recent shape completion methods can offer an alternative by reconstructing spinal structures in US data, while being pretrained on large set of publicly available CT scans. However, these approaches are typically offline and have limited reproducibility.  
In this work, we introduce a novel integrated system that combines robotic ultrasound with real-time shape completion to enhance spinal visualization. Our robotic platform autonomously acquires US sweeps of the lumbar spine, extracts vertebral surfaces from ultrasound, and reconstructs the complete anatomy using a deep learning-based shape completion network.
This framework provides interactive, real-time visualization with the capability to autonomously repeat scans and can enable navigation to target locations. This can contribute to better  consistency, reproducibility, and understanding of the underlying anatomy. We validate our approach through quantitative experiments assessing shape completion accuracy and evaluations of multiple spine acquisition protocols on a phantom setup. Additionally, we present qualitative results of the visualization on a volunteer scan.

\end{abstract}

\section{INTRODUCTION}
\begin{figure}[t]
    \centering
    \includegraphics[width=\columnwidth, keepaspectratio]{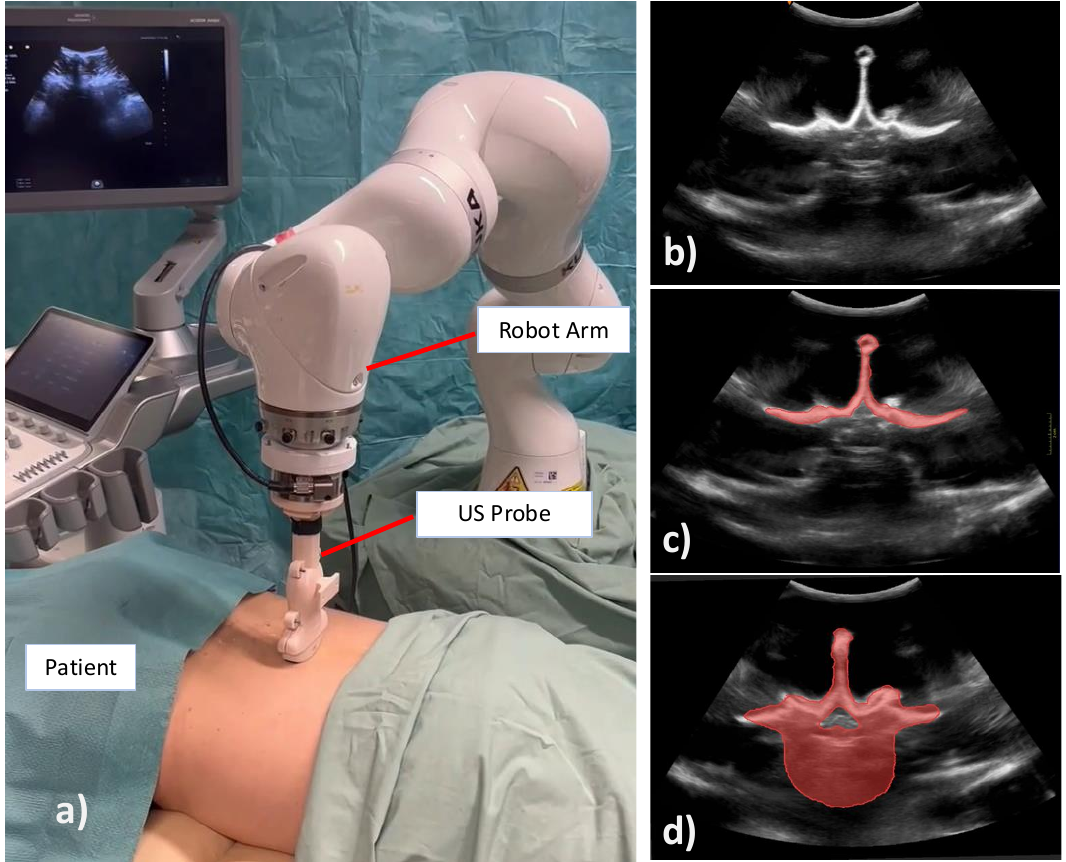}
    \caption{Overview of the proposed system: a) we use a robotic arm with an attached convex ultrasound probe to scan the patient b) Live B-Mode ulttrasound image. c) Segmentation mask of bone d) Shape completion of the full vertebral anatomy.}
    \label{Fig:teaser_fig}
\end{figure}

Ultrasound imaging has increasingly become as a valuable tool in various spinal procedures, largely due to its real-time, radiation-free capabilities that significantly enhance clinical precision and safety. 
In spinal injections, such as facet joint and epidural injections, US provides guidance to improve needle placement accuracy while reducing radiation exposure compared to fluoroscopy \cite{rasoulian2015ultrasound}. In surgical settings, such as spinal decompression, intraoperative US facilitates the continuous assessment of spinal cord pulsations and dural movements, thereby allowing surgeons to fine-tune decompression techniques based on real-time feedback \cite{kimura2012ultrasonographic}. In the context of spinal tumor resections,  US plays a critical role in delineating tumor margins intraoperatively, which aids in achieving precise resection and minimizes the risk of leaving residual tumor tissue \cite{zhou2011intraoperative}. Additionally, US has been explored for lumbar punctures, where it assists in identifying optimal needle insertion points, particularly in patients with challenging anatomical variations \cite{xu2022autoinfocus}. Overall, the integration of US in spinal interventions not only provides real-time imaging and avoids ionizing radiation but also enhances procedural accuracy and intraoperative decision-making, making it a versatile and effective tool for spinal interventions.

In spite of these advantages, US guidance remains challenging in spine procedures due to shadowing artifacts caused by the surface of the spine, that prevents the US beam from penetrating deeper tissue, making it difficult to accurately localize the target of interest \cite{li2021image}.
Traditionally, the lack of sufficient anatomic information for guidance in US image has been tackled through CT/US registration, which brings anatomical information from a CT scan into real-time operating room scenario. 
However, CT-to-US spine registration is inherently complex, as it requires anatomical knowledge about vertebral relationships and is limited by differences in spine curvature between modalities, as patient positioning changes between pre-operative and intra-operative stages \cite{azampour2024anatomy,nagpal2015multi}. Furthermore, this approach necessitates recent diagnostic CT scan, which is not always available for every patient.

\begin{figure*}[th]
    \centering
    \includegraphics[width=\textwidth, keepaspectratio]{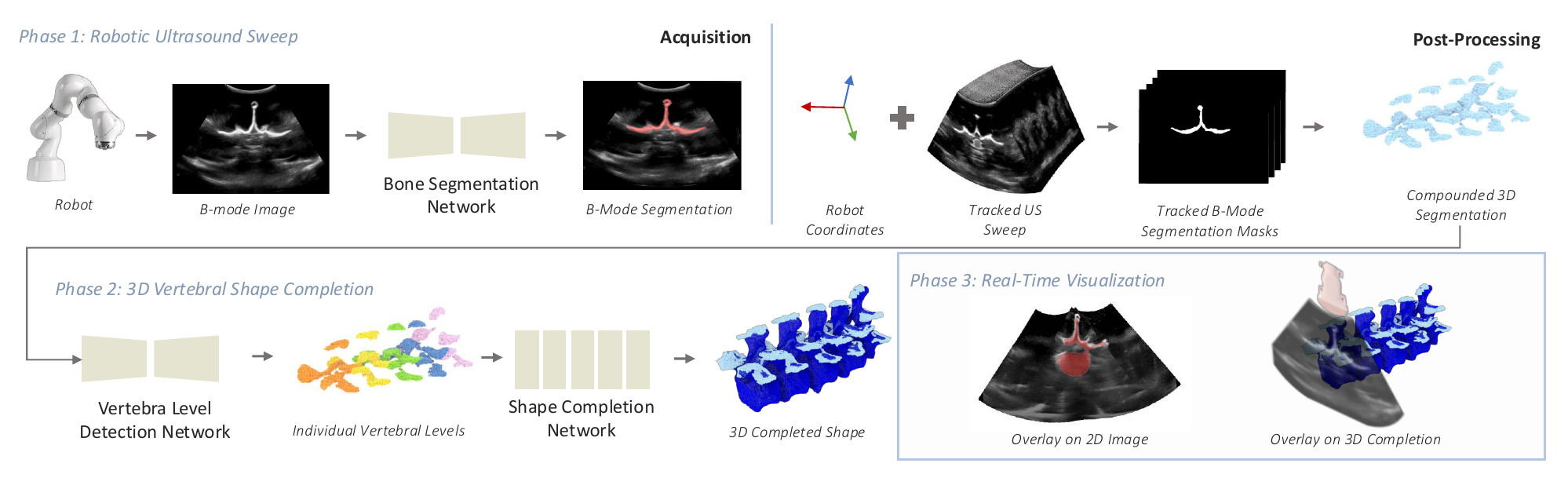}
    \caption{The proposed robotic ultrasound imaging system for vertebral reconstruction consists of three phases: (1) automatic ultrasound acquisitions to capture and segment bone structures creating 3D segmentations, (2) processing of B-Mode images to generate a complete 3D spine model from partial vertebrae observations, and (3) real-time visualization overlaying reconstructed vertebral models onto 2D ultrasound images and 3D renderings as the robot follows the same trajectory.}
    \label{Fig:methodology}
\end{figure*}

Recent work by Gafencu et al.~\cite{gafencu2024shape} has proposed an alternative approach using shape completion techniques to reconstruct missing spine structures in ultrasound, thereby alleviating the need for CT-to-US registration. By learning the distribution of the complete vertebral shapes during training, this method demonstrates realistic completions both in phantom and patient scans. 
During inference the method requires partial point cloud of the vertebral surfaces obtained from ultrasound sweeps. However, this is an offline approach,
lacking real-time capabilities, making repeatability and adjustments of the scan not possible. 
Moreover, the quality of the completion is dependent on the acquired scan, which is heavily dependent on the operator and their experience level.
\par

To enable automatization of the acquisition, robotic ultrasound systems for spine procedures have garnered significant attention due to their potential to deliver automated, consistent, and reproducible imaging — a crucial factor for enhancing clinical outcomes. Recent works have showcased how these systems can leverage force feedback and advanced control algorithms to accurately position the US probe, ensuring stable contact and optimal image quality \cite{tirindelli2020force}. Further approaches employ deep reinforcement learning to enable autonomous probe navigation, dynamically adjusting trajectories to maintain ideal imaging conditions throughout the procedure \cite{hase2020ultrasound}. These developments reduce the dependence on operator skill and variability, leading to more reliable imaging and better-informed clinical decisions. 

\par

In this work, we introduce a novel integrated system combining robotic ultrasound with shape completion for enhanced spine visualization (Figure \ref{Fig:teaser_fig}). Our robotic system autonomously acquires ultrasound sweep of the lumbar spine, followed by extracting of the vertebral surfaces and reconstructing missing anatomy using a deep learning-based shape completion network trained solely on publicly available CT data. 
Our framework provides interactive, real-time visualization, wherein the completed vertebral structures can be overlaid on live US images, and the robot can autonomously repeat scans. This can enable navigation to clinician-defined target locations for improved guidance.
We perform both quantitative experiments on the accuracy of the shape completion and evaluation of multiple spine acquisition protocols in both phantom and patient setup.
Additionally, we present qualitative results of the real time visualization on a volunteer scan. 
By inferring underlying spinal structures in real time, our approach can enhance procedural consistency, reproducibility, and intraoperative decision-making. The code can be found at https://github.com/ReemShaban/roboUS-completion.

\section{METHODOLOGY}

Our proposed system for autonomous ultrasound-guided lumbar spine visualization combines robotic ultrasound acquisition with real-time vertebral shape completion. The methodology consists of three main phases: (1) Robotic Ultrasound Sweep (2) 3D Vertebral Shape Completion (3) Real-time Visualization, as illustrated in Fig.~\ref{Fig:methodology}.

\subsection{Robotic Ultrasound Sweep}
\subsubsection{Robotic setup}
We employ a 7-DOF robotic arm with a force-torque sensor mounted at the end-effector. An ultrasound probe is attached to the robot's end-effector using a custom 3D-printed holder. The system is calibrated following the cone calibration proposed by Ronchetti et. al.~\cite{ronchetti2022pro} to establish the transformation between the robot base, ultrasound probe, and image coordinate systems.
The calibration process determines the transformation matrix $T_{probe}^{ee}$ that relates the ultrasound image coordinates to the robot's end-effector frame:

\begin{equation}
T_{probe}^{ee} = \begin{bmatrix}
R_{probe}^{ee} & t_{probe}^{ee} \\
0 & 1
\end{bmatrix}
\end{equation}

where $R_{probe}^{ee} \in \mathbb{R}^{3 \times 3}$ is the rotation matrix and $t_{probe}^{ee} \in \mathbb{R}^{3}$ is the translation vector. This transformation allows us to position each ultrasound pixel in 3D space using the formula:

\begin{equation}
\mathbf{p}_{world} = T_{base}^{world} \cdot T_{ee}^{base} \cdot T_{probe}^{ee} \cdot T_{image}^{probe} \cdot \mathbf{p}_{image}
\end{equation}

\begin{figure}[t]
    \centering
    \includegraphics[width=\columnwidth, keepaspectratio]{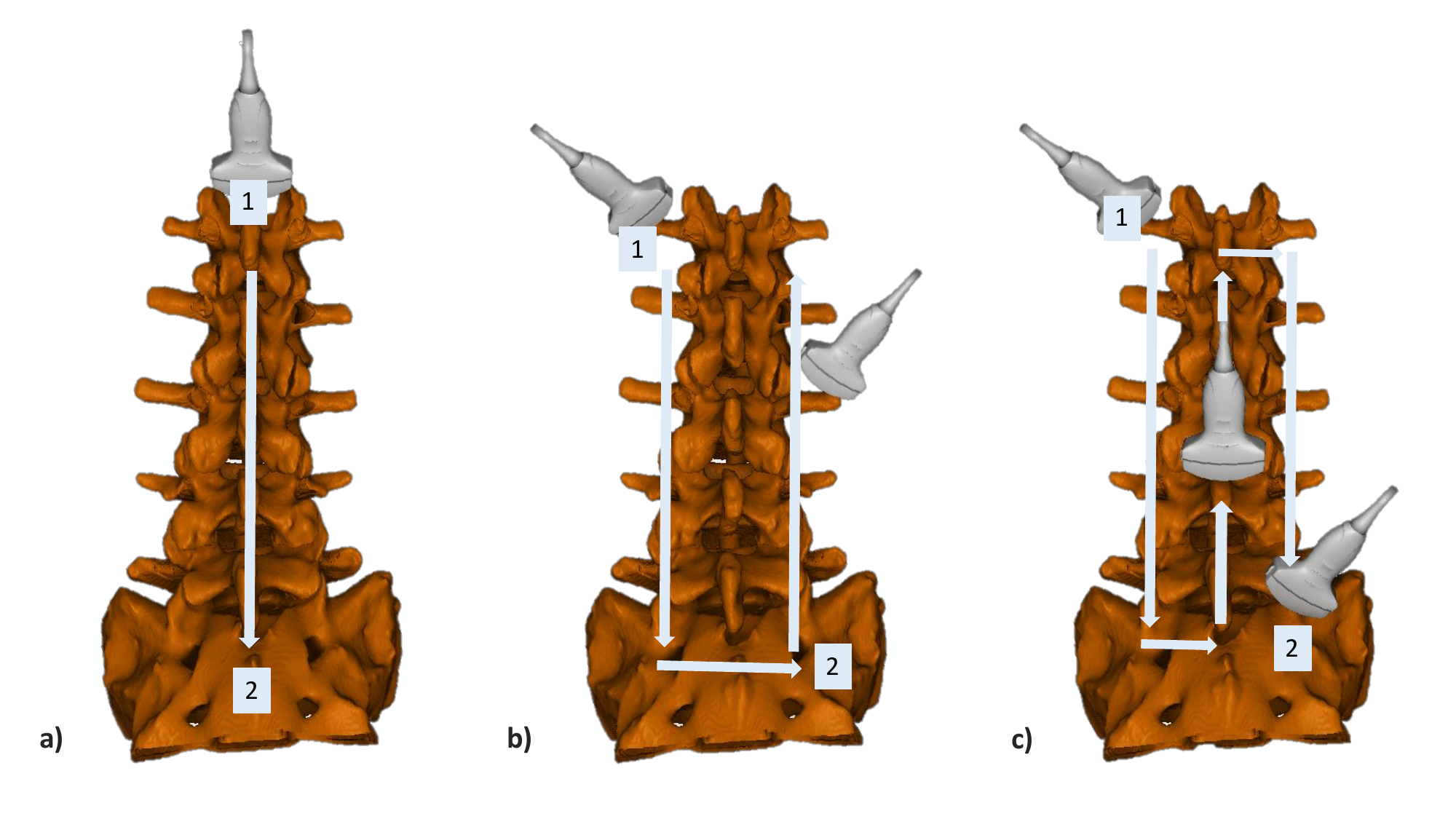}
    \caption{Three types of automatic scans evaluated for optimal shape completion results: a) Linear scan, b) U-shape scan, c) Zig-Zag scan. \revisionAdd{Two points are manually initialized for computing the trajectories.}}
    \label{Fig:acquisition}
\end{figure}

\subsubsection{Image acquisition and scanning trajectories} 
\label{subsec:scantraj}
We designed three robotic path planning strategies for acquisition: Linear, U-shape, and Zig-Zag scans, illustrated in Figure \ref{Fig:acquisition}. Due to patient-specific variations in lumbar anatomy, we require manual input at the beginning of each scan, followed by a fully automated acquisition process.

For the linear scan, the operator initially positions the ultrasound probe transversely at the intervertebral space between vertebrae T12 and L1, ensuring symmetric visualization of transverse processes. This initial point ensures full coverage of L1. The operator then sets the final position at the sacral joint, maintaining the same probe orientation. The robot then navigates in a linear trajectory with an update in position along the longitudinal axis of the spine.
In the U-shape scan, the probe is placed transversely centered on the lateral process of L1. The operator then sets a diagonally opposite endpoint as indicated in Figure \ref{Fig:acquisition}. The robot navigates in three segments, two longitudinal and one transverse, with constant positional updates.
The Zig-Zag scan combines elements of the linear and U-shape trajectories. Similar to the U-shape scan, the operator defines start and end points on opposite sides of the transverse processes of L1 and L5. The robot navigates in five segments, three longitudinal and two transverse, to comprehensively cover the spine’s central and lateral regions.
During all acquisitions, the robot operates in force-controlled mode, ensuring consistent probe-skin contact for optimal imaging quality.

\subsubsection{Automatic B-Mode image segmentation}
To segment the surface of vertebrae in ultrasound images, we train a 2D U-Net architecture, with four encoder-decoder layers and skip connections to effectively capture both local and global features. Our training dataset consists of manually acquired sparse set of phantom images, collected with freehand scanning. Unlike automatic scans, manual acquisitions are operator-dependent, introducing variability in probe positioning, pressure, angle, and speed. This results in greater diversity in image quality and anatomical representation, which encourages the network to be more robust and adaptable across different settings. At inference time, the network processes B-mode ultrasound images obtained from automatic robotic acquisitions and outputs a binary segmentation mask highlighting the visible vertebral structures.

\subsection{3D Vertebral Shape Completion}

\subsubsection{3D Compounding and vertebral level identification}
\label{subsec:pointnet}
To utilize shape completion to recover missing information in the ultrasound scan, we require individual observations from each vertebral level. First, we compound the segmentation masks into a 3D volume, using their 3D poses derived from the tracked robot coordinates during the acquisition. Depending on the acquisition protocol, the compounding can be done directly (for Linear scan) or requires an additional step, where for the values where two images overlap the maximum is taken (for U-shape and Zig-Zag scan).
The 3D compounded segmentation is first converted from a label map to a mesh using the marching cubes algorithm. The vertices of the mesh are sampled via furthest point sampling to obtain the point cloud of the spine surface  \(\mathcal{S}\):

\begin{equation}
\mathcal{S} = \{\mathbf{p}_i \in \mathbb{R}^3 \mid i = 1, 2, \dots, N\}
\end{equation}

where \(N\) represents the number of sampled points.

\noindent
To classify each point into its corresponding vertebral level, we train a PointNet~\cite{qi2017pointnet} to assign each point \(\mathbf{p}_i\) in the spine surface \(\mathcal{S}\) to one of the five lumbar vertebrae (L1–L5). Each point is thus associated with a predicted class label \( \hat{y}_i \):
\begin{equation}
\hat{y}_i \in \{1,2,3,4,5\}, \quad \forall \mathbf{p}_i \in \mathcal{S}, \quad i = 1, 2, \dots, N.
\end{equation}
For training, we use the large-scale, annotated VerSe20~\cite{sekuboyina2021verse} dataset of computed tomography (CT) scans. Following prior work~\cite{gafencu2024shape}, we employ ultrasound-aware ray casting to extract the spine surface, mimicking ultrasound acquisition. Given vertebra-wise annotations, we can accurately associate each surface point with its corresponding vertebra.
At inference, the trained PointNet model processes the segmented spine surface. It passes the input through a series of shared multilayer perceptrons (MLPs) and outputs the predicted vertebral labels for each point. 

\subsubsection{Shape completion}\label{subsub:shape_completion}
Each identified vertebral level is run through the shape completion model to reconstruct missing regions. To provide sufficient context for this process, we use a bounding box approach that includes partial observations of neighboring vertebrae.
For shape completion, we adopt the framework by Gafencu et al.~\cite{gafencu2024shape}, which learns to recover complete vertebral shapes using pairs of incomplete and complete vertebrae.
We use the ray-casted spine meshes generated in \ref{subsec:pointnet} for training. The method employs two networks trained end-to-end: one performs coarse shape completion by learning the prior distribution of vertebral shapes, while the other refines the output to preserve details.
The network is trained using a joint loss:
\begin{equation} \mathcal{L} = \lambda_{\text{CD}} \cdot \mathcal{L}{\text{CD}} + \lambda{\text{KL}} \cdot \mathcal{L}_{\text{KL}}. \end{equation}
The Chamfer Distance term ensures proximity between the predicted and ground truth point clouds, while the Kullback-Leibler (KL) divergence encourages the partial shape distribution to match the complete shape distribution, which helps the model learn a meaningful latent space that can generate complete shapes from partial observations.
During inference, each identified vertebral level, along with additional points from the bounding box, is processed through the shape completion network.

\subsection{Real-time Visualization}

Firstly, we acquire one complete scan of the spine with the selected scanning trajectory from \ref{subsec:scantraj}
 and pass the acquired images to the 3D vertebral shape completion pipeline. 
Subsequently, to enable real-time visualization, the robot executes the same trajectory again while visualizing the overlaid completion on the real-time ultrasound image, enabling the clinician to see the augmented view simultaneously.


\begin{figure}[t]
    \centering
    \includegraphics[width=0.8\columnwidth, keepaspectratio]{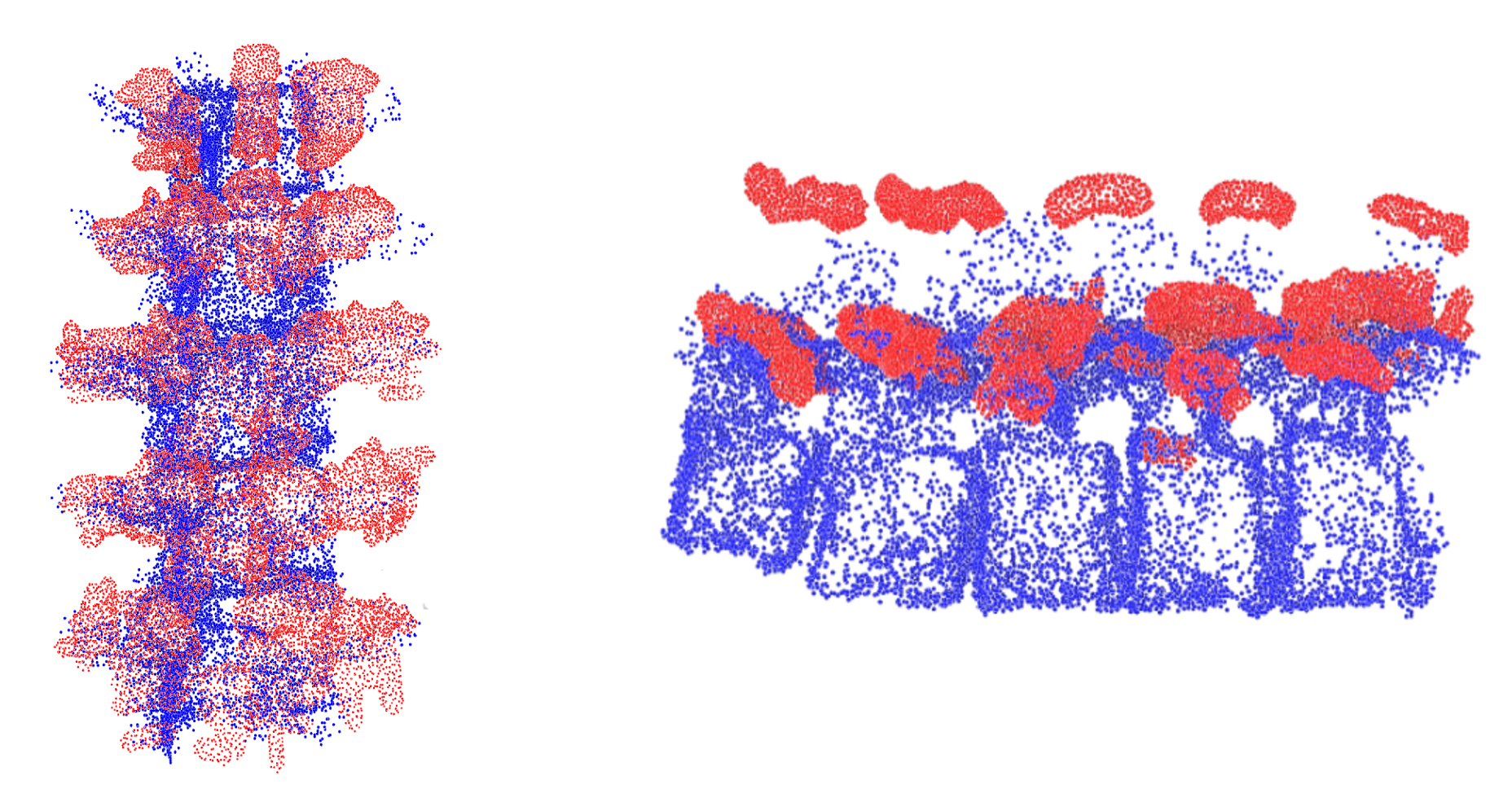}
    \caption{Qualitative results of our proposed robotic imaging system on the volunteer data are presented, showcasing two views of the shape completion outcomes on volunteer datasets. The red pointcloud represent the partial point cloud obtained from ultrasound, while the blue regions depict the completed shape. }
    \label{Fig:vol_qual}
\end{figure}

\section{SETUP}
1) \textit{Hardware}
An ACUSON Juniper 5C1 curvilinear probe\footnote{Siemens Healthineers, Germany} was mounted on a KUKA LBR iiwa 14 R820\footnote{KUKA LBR iiwa 14 R820, KUKA Roboter GmbH, Augsburg, Germany} using a 3D-printed holder. A constant force of 5N was applied at 500 N/m stiffness. Control was handled via the iiwa stack and ROS, with images acquired by a frame grabber\footnote{Epiphan Video, Canada} and streamed live through ImFusion Suite\footnote{ImFusion GmbH, Germany} at a 13 cm depth.

2) \textit{Training Details}\label{setup_training}
We train the shape completion network for 100 epochs with a best-network policy using an Adam optimizer and learning rate or 0.0001. At training the batch size is set to 4, while at inference we set it to 2. 

3) \textit{Evaluation Metrics}
We utilize three different shape completion metrics to evaluate the accuracy of our proposed pipeline: Chamfer Distance (CD), Earth Mover's Distance (EMD) and F1 score. 
Chamfer Distance, measures the point-to-point distance between the completed shape and the ground truth. EMD quantifies shape dissimilarity by calculating the minimum effort needed to transform one shape into another. For better interpretability, we scale our CD and EMD values by $10^4$.
To handle outliers effectively, we utilize F1-score. This metric combines precision and recall as a harmonic mean, as an additional performance indicator.

\begin{figure}[t!]
    \centering
    \includegraphics[width=0.8\columnwidth, keepaspectratio]{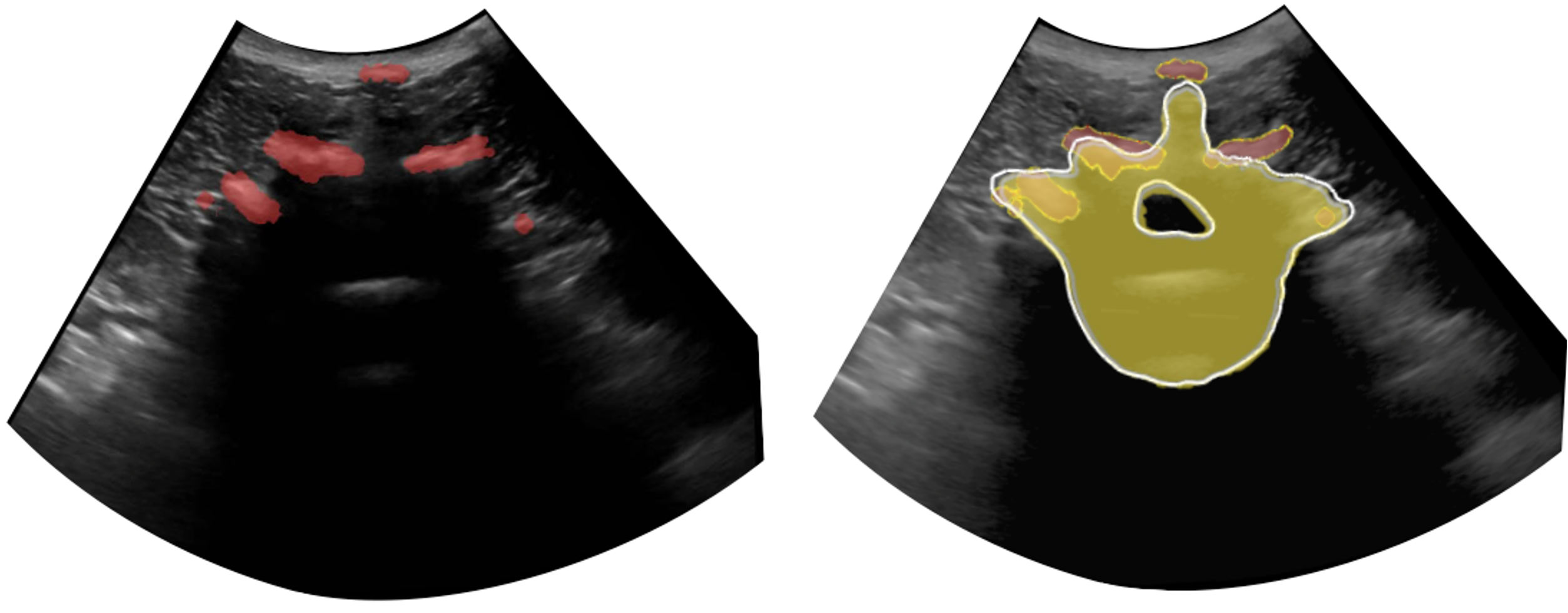}
    \caption{Qualitative results of our proposed robotic imaging system are shown on the volunteer data. On the left, the live B-mode image is displayed, while on the right, the overlaid completed shape. }
    \label{Fig:vol_qual2}
\end{figure}

\section{EXPERIMENTS}
To validate our robotic ultrasound framework, we perform three main experiments detailed in the following.

\subsection{Evaluation of scanning trajectories for shape completion}

An evaluation of the shape completion performance was conducted across the three robotic ultrasound acquisition methods depicted in Figure \ref{Fig:acquisition}. To enable quantitative assessment, a computed tomography (CT) scan of the phantom was acquired and utilized as the ground truth for comparison.
Ultrasound imaging of the spine presents significant challenges due to the presence of complex anatomical structures and artifacts induced by bone surfaces, which can obscure key morphological details. To mitigate these limitations and enhance imaging quality, three distinct scanning trajectories were implemented, with the objective to assess how variations in scanning patterns influence image quality, anatomical coverage, and overall efficacy in spinal ultrasound acquisition for shape completion. Each trajectory was executed under standardized conditions, ensuring uniform imaging parameters, including transducer frequency and depth settings. The acquired B-mode ultrasound images were subsequently processed through our structured pipeline, including U-Net for vertebral bones segmentation, followed by PointNet individual vertebra level segmentation, and a final shape completion network for comparative analysis. The results from this experiment can offer insights into the optimal scanning trajectory for spinal ultrasound, with implications for improving accuracy and enhancing image-guided spinal interventions.

\subsection{Evaluation of the complete pipeline on a volunteer}
We acquire robotic ultrasound scans of a volunteer using a linear scanning approach, followed by manual segmentation. The goal of this experiment was to assess the performance of our pipeline on real human data, as opposed to phantom-based evaluation alone. While phantoms provide a controlled environment for testing, they do not fully capture the complexities encountered in real human scanning.
In particular, factors such as tissue heterogeneity, motion artifacts, and variations in acoustic properties introduce additional complexities not present in synthetic or phantom data. Furthermore, the compounding process in human subjects is more challenging due to variations in probe contact, alignment inconsistencies, and anatomical differences. By performing this evaluation, we assess our approach robustness to real-world conditions and applicability for clinical use.
\subsection{Patient-specific refinement}
Accurate whole-spine reconstruction is critical for ensuring precise surgical guidance in spine procedures. However, ultrasound imaging inherently lacks visibility of features like the vertebral bodies, leading to significant ambiguity in shape and size reconstruction. This limitation can introduce errors that compromise the accuracy of intraoperative navigation. 
\revisionAdd{To address this, we investigate how incorporating preoperative imaging data can enhance vertebral shape completion. For two patients with an available preoperative CT scan, we use the US-aware raycasting method from Section~\ref{subsec:pointnet} to simulate ultrasound-consistent spine surfaces. Paired with complete shapes extracted from CT, they are used to fine-tune the pretrained shape completion network for 10 additional epochs, following the training procedure described in Section~\ref{setup_training}. This allows the network to learn patient-specific anatomical details. To evaluate the benefit of this approach, we compare spine reconstructions from real US acquisitions of these two patients using the population-based prior network described in Section~\ref{subsub:shape_completion} and the patient-specific fine-tuned network.}


\begin{figure}[t]
    \centering
    \includegraphics[width=\columnwidth, keepaspectratio]{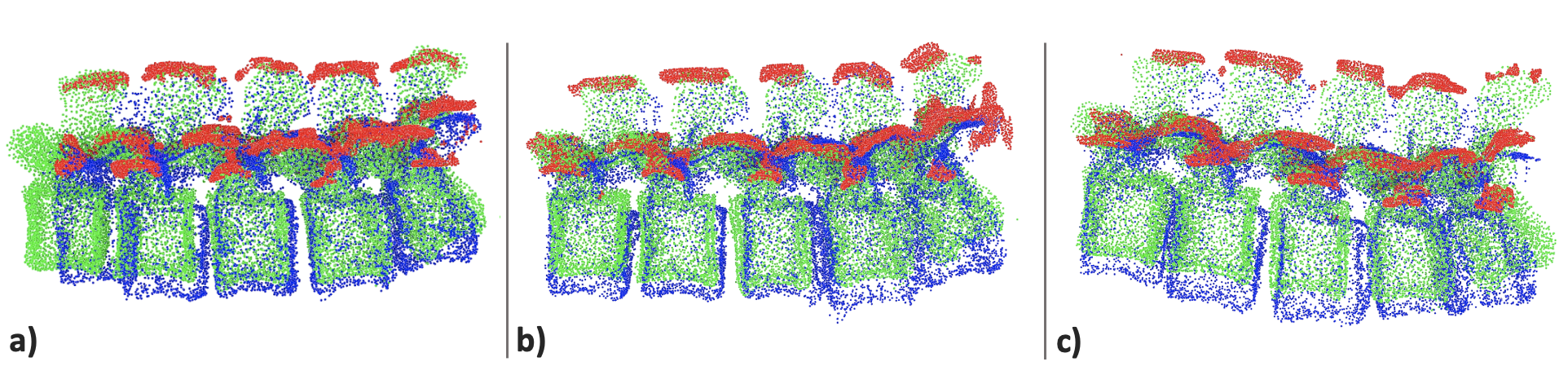}
    \caption{Qualitative results of the phantom shape completion for the three acquisition methods: linear (a), U-shape (b) and Zig-Zag (c) displaying the input(red), completion(blue) and ground truth (green), obtained through segmenting the CT scan}
    \label{Fig:vol_qual3}
\end{figure}
\section{RESULTS \& DISCUSSION}

\subsection{Comparison of acquisition trajectories}

The results in Table \ref{table:shapecomplmetrics} compare the three acquisition strategies: Linear, U-shape, and Zig-Zag based. The linear scan generally outperforms the others, particularly in CD and EMD, which measure the geometric accuracy of the reconstructed point clouds. However, an anomaly is observed in the L1 vertebra, where the CD for the linear scan is significantly higher, likely due to an incomplete segmentation. Without this outlier, the mean CD for the linear scan drops from 20.30 to 12.71. 
A lower CD ensures better preservation of local anatomical structures, crucial for procedures such as surgical navigation and implant placement. 
Inaccurate reconstructions can lead to misalignments, making local feature preservation a key objective. 
While CD shows clear differences, the other two metrics exhibit minimal variations. EMD values differ by only 0.01, indicating all methods capture the global shape similarly well. F1 scores vary by just 0.02, suggesting all approaches reconstruct a comparable amount of anatomy. This highlights CD as the most sensitive metric for evaluating fine structural details, where the linear scan excels. Despite the L1 segmentation issue, the linear scan remains the most reliable approach for completing the shape of a spine phantom, ensuring both local detail preservation and global shape accuracy. 

While, the U-shape and Zig-Zag acquisitions  cover a larger surface area compared to the linear method, they do not consistently produce better results. This is likely due to the compounding process, where overlaps or misalignments between consecutive sweeps may decrease the cohesiveness of the resulting volume and introduce noise or artifacts. A prominent observation across all acquisition methods is the poorer performance at the furthest vertebral levels, particularly at L5, which is near the sacrum and may be less clearly captured in the ultrasound sweeps. 

\begin{table*}[h]
    \caption{Shape completion metrics comparing the three ultrasound acquisition methods}
    \centering
    \label{table:shapecomplmetrics}
    \resizebox{\textwidth}{!}{
    \begin{tabular}{lccc|ccc|ccc}   
        \toprule
        \textbf{Vertebra} & \multicolumn{3}{c|}{\textbf{CD$\downarrow$}} & \multicolumn{3}{c|}{\textbf{EMD$\downarrow$}} & \multicolumn{3}{c}{\textbf{F1$\uparrow$}} \\
        & \textbf{Linear} & \textbf{U-shape} &  \textbf{Zig-Zag}  & \textbf{Linear}  & \textbf{U-shape}  & \textbf{Zig-Zag} & \textbf{Linear} & \textbf{U-shape} & \textbf{Zig-Zag} \\
        \midrule
        L1  & 50.63 & 19.98  & \textbf{14.73}  & 0.14  & \textbf{0.09}  & 0.10 & 0.11 & 0.17 & \textbf{0.20}  \\
        L2  & 14.07 & \textbf{11.55}  & 15.69 & 0.09  & \textbf{0.08}  & \textbf{0.08} & 0.19 & 0.19 & \textbf{0.22}   \\
        L3  & \textbf{8.50 }& 14.24  & 19.62  & \textbf{0.07}  & 0.10  & 0.08 & \textbf{0.24} & 0.17 & 0.17   \\
        L4  & \textbf{9.44} & 13.81  & 13.76  & 0.09  & \textbf{0.07}  & \textbf{0.07} & 0.20 & 0.14 & \textbf{0.22}   \\
        L5  & \textbf{18.85} & 23.72  & 36.06  & \textbf{0.10}  & 0.13  & \textbf{0.10} & \textbf{0.17} & 0.15 & 0.13   \\
        \midrule
        \textbf{Average} & 20.01 ± 15.61 & \textbf{17.84 ± 4.49} & 22.65 ± 8.29 & 0.10 ± 0.02 & 0.10 ± 0.02  & \textbf{0.09 ± 0.01} & 0.18 ± 0.04 & \textbf{0.16 ± 0.01} & 0.18 ± 0.03 \\
        \bottomrule
    \end{tabular}
    }
\end{table*}

\subsection{Qualitative evaluation of shape completion on volunteer}
We acquire robotic ultrasound scans of a volunteer using the linear scanning approach, followed by manual segmentation. Since no ground truth is available (e.g. from a pre-operatice scan), we present the qualitative results in Figures \ref{Fig:vol_qual} and \ref{Fig:vol_qual2}. We observe that the lateral processes in the predicted completion align well with the input, but discrepancies exist in the spinous processes. Vertebral bodies match the expected size, which can be estimated from the dimensions of the vertebral arches individually. 
Furthermore, as segmentation of vertebrae in ultrasound is a challenging task, errors in the segmentation can additionally contribute to inconsistencies in the completion results. However, using our proposed robotic ultrasound imaging acquisition system on a volunteer provides with realistic shape completion. These preliminary results indicate the potential of our pipeline to be integrated in spinal ultrasound interventions. 
Additionally, within a clinical scenario, combining 3D visualization as shown in Figure \ref{Fig:vol_qual} with slice-by-slice analysis depicted in Figure \ref{Fig:vol_qual2} can reduce the cognitive burden on clinicians, eliminating the need for mental reconstruction of 3D anatomy. Together with 3D visualization of the spine, it can serve as a reference for visualizing critical structures such as nerves and may prove useful in the planning phase.

\subsection{Patient-specific shape completion}

Table \ref{table:tablepatientspecif} compares shape completion trained only on ultrasound versus fine-tuning on patient-specific CT, showing consistent improvement across all metrics. CD improved by an average of 4.93, with the largest gain at L5 (from 20.42 to 6.02), highlighting the benefit of prior shape knowledge. EMD showed modest but consistent reductions, on average 4.58\%, peaking at L3 (10.00\%). 
The F1 score showed the greatest improvement (3.37 times increase on average), with peaks of 5 times at L5, 4 times at L3, and 3.36 times at L2.

These improvements underscore the effectiveness of integrating preoperative CT priors, to reduce shape ambiguity, for more precise vertebral reconstruction.
Additionally, as refinement occurs pre-operatively, the approach remains computationally efficient for real-time intra-operative use, making it a suitable for real-time visualization during robotic-assisted procedures. 
\begin{table}[h]
    \caption{Shape completion metrics comparing ultrasound-based method (US) with patient specific refinement (Ref.).}
    \centering
    \begin{tabular}{lcc|cc|cc}
        \toprule
        \textbf{Vertebra} & \multicolumn{2}{c|}{\textbf{CD$\downarrow$}} & \multicolumn{2}{c|}{\textbf{EMD$\downarrow$}} & \multicolumn{2}{c}{\textbf{F1$\uparrow$}} \\
        & \textbf{Ref.} & \textbf{US} & \textbf{Ref.} & \textbf{US} & \textbf{Ref.} & \textbf{US} \\
        \midrule
        L2  & \textbf{5.54}  & 9.18  & \textbf{0.10}  & 0.10  & \textbf{0.37}  & 0.11  \\
        L3  & \textbf{9.36}  & 11.34 & \textbf{0.09}  & 0.10  & \textbf{0.32}  & 0.08  \\
        L4  & \textbf{5.84}  & 5.54  & \textbf{0.05}  & 0.05  & \textbf{0.11}  & 0.10  \\
        L5  & \textbf{6.02}  & 20.42 & \textbf{0.11}  & 0.12  & \textbf{0.30}  & 0.06  \\
        \midrule
        \textbf{Average} & \textbf{6.69} & 11.62 & \textbf{0.09} & 0.09 & \textbf{0.27} & 0.08 \\
        \bottomrule
    \end{tabular}
    \label{table:tablepatientspecif}
\end{table}

\section*{CONCLUSION \& FUTURE WORK}
We present an integrated system combining robotic ultrasound with shape completion to enhance spinal visualization. The robotic platform autonomously acquires lumbar spine US sweeps, extracts vertebral surfaces, and reconstructs the complete anatomy using a shape completion network. Quantitative and qualitative evaluations on both phantom and patient studies demonstrate the accuracy of our shape completion and the robustness of different spine acquisition protocols, while a volunteer scan showcases the feasibility of real-time visualization.
\revisionAdd{While initial feasibility has been demonstrated, future work should focus on optimizing the robotic acquisition trajectory using contextual information from the evolving shape estimate. In addition, reconstruction accuracy could be improved by leveraging relative pose information across adjacent vertebrae and completing the lumbar spine as a whole, rather than one vertebra at a time. These enhancements would not only improve reconstruction accuracy, but also increase the system’s adaptability to different anatomical scenarios and patient-specific variations. Overall, integrating robotic ultrasound with real-time shape completion holds strong potential to advance spinal imaging and enable more autonomous and reliable interventions.
}


\section*{ACKNOWLEDGMENT}
The work was supported in part by the Multi-Scale Medical Robotics Center, AIR@InnoHK, Hong Kong; and in part by the SINO-German Mobility Project under Grant M0221.

\bibliographystyle{unsrt}
\bibliography{refs}

\end{document}